\documentclass[acmtog]{acmart}

\usepackage{booktabs} 

\usepackage{amsmath}
\usepackage{nicefrac}
\usepackage[ruled]{algorithm2e} 

\SetAlFnt{\small}
\SetAlCapFnt{\small}
\SetAlCapNameFnt{\small}
\SetAlCapHSkip{0pt}
\IncMargin{-\parindent}

\usepackage{dblfloatfix}
\usepackage{float}

\usepackage{subcaption}

\copyrightyear{2026}
\acmYear{2026}
\setcopyright{cc}
\setcctype{by}
\acmConference[SIGGRAPH Conference Papers '26]{Special Interest Group on Computer Graphics and Interactive Techniques Conference Conference Papers}{July 19--23, 2026}{Los Angeles, CA, USA}
\acmBooktitle{Special Interest Group on Computer Graphics and Interactive Techniques Conference Conference Papers (SIGGRAPH Conference Papers '26), July 19--23, 2026, Los Angeles, CA, USA}
\acmDOI{10.1145/3799902.3811086}
\acmISBN{979-8-4007-2554-8/2026/07}

\begin{document}

\title{Broadband Hyperspectral 3D Imaging using Dispersed Structured Light}

\author{Suhyun Shin}
\email{}
\affiliation{%
  \institution{POSTECH}
  \country{South Korea}
}

\author{Yunseong Moon}
\email{}
\affiliation{%
  \institution{POSTECH}
  \country{South Korea}
}

\author{Ryota Maeda}
\email{}
\affiliation{%
  \institution{University of Hyogo}
  \country{Japan}
}

\author{David B. Lindell}
\email{lindell@cs.toronto.edu}
\affiliation{%
  \institution{University of Toronto}
  \country{Canada}
}


\author{Kiriakos N. Kutulakos}
\email{kyros@cs.toronto.edu}
\affiliation{%
  \institution{University of Toronto}
  \country{Canada}
}

\author{Seung-Hwan Baek}
\email{shwbaek@postech.ac.kr}
\affiliation{%
  \institution{POSTECH}
  \country{South Korea}
}

\authorsaddresses{}

\begin{abstract}

Hyperspectral 3D imaging enables the capture of dense spectral information and scene geometry but has traditionally been confined to narrow spectral windows, typically the visible range. In this work, we introduce a broadband hyperspectral 3D imaging (BH3D) method to extend this capability across the full visible-near-infrared and short-wavelength infrared (SWIR) spectrum (450–1500\,nm). This broad coverage is critical as it captures complementary physical cues: visible wavelengths reveal surface appearance, while SWIR bands provide insight into subsurface properties and material composition. However, realizing BH3D is challenging due to fundamental sensor constraints between visible-spectrum silicon and SWIR-spectrum InGaAs sensors, which necessitate complex multi-spectrograph designs. Here we propose a single-spectrograph BH3D system, using a stereo setup comprising visible and SWIR cameras, that reconstructs dense broadband hyperspectral reflectance together with accurate 3D geometry. Our key idea is to extend dispersed structured light to the broadband regime using a single spectrograph. We model the image formation of broadband dispersed structured light, and estimate hyperspectral reflectance and depth. We validate our approach on diverse real-world scenes, demonstrating accurate reconstruction with a mean spectral angle mapper of 0.13 rad, root mean square error of 0.03, and mean depth error of 4.5\,mm. We further demonstrate identifying metameric materials, performing imaging through opaque layers, uncovering hidden features on banknotes, and revealing blood vessels.

\end{abstract}

\begin{teaserfigure}
  \includegraphics[width=\linewidth]{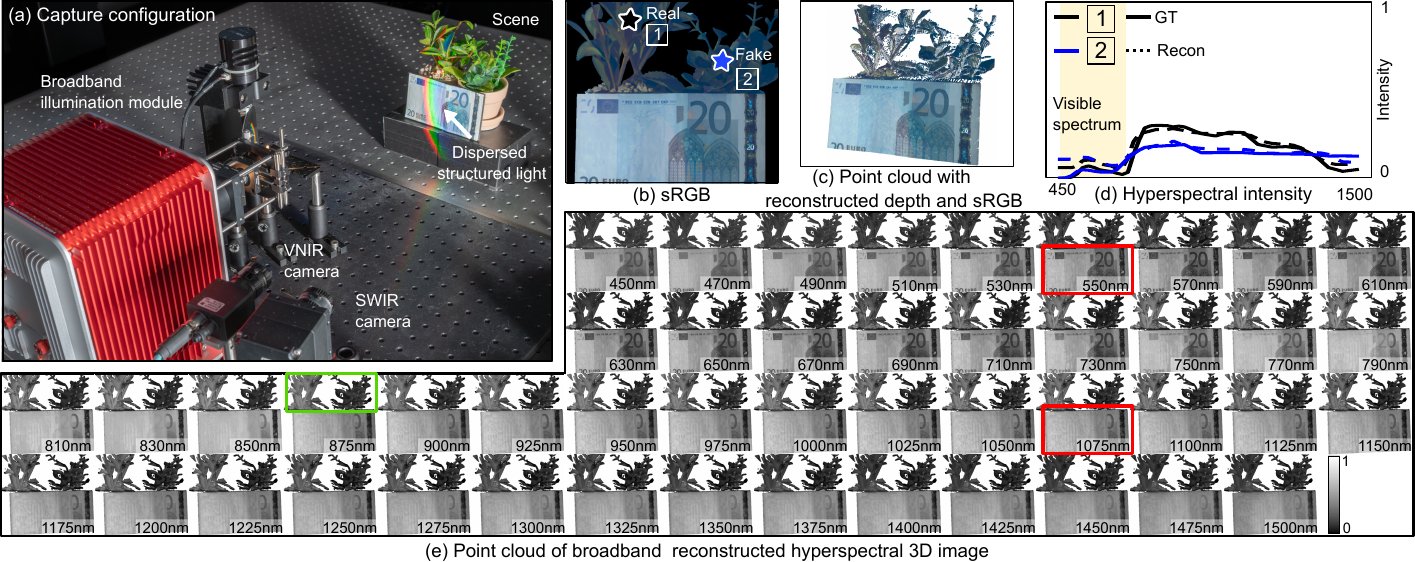}
  \caption{We demonstrate broadband hyperspectral 3D imaging (BH3D) from visible to short-wavelength infrared spectrum (450--1500\,nm) with dispersed structured light. (a) Capture configuration, (b) reconstructed hyperspectral image in sRGB and (c) point cloud with reconstructed depth and sRGB, (d) hyperspectral reflectances for the real plant and the fake plant. (e) Point cloud of reconstructed broadband hyperspectral reflectance, showing security patterns on the banknote and clear spectral differences between real and artificial plants. Red and green boxes highlight the bill patterns and plant differences, respectively.}
  \label{fig:teaser}
\end{teaserfigure}


\maketitle


\newcommand{\vect}[1]{\mathbf{#1}}
\newcommand{\mat}[1]{\mathbf{#1}}

\newcommand\note[1]{\textcolor{red}{#1}}

\newcommand{\psf}{\rho}

\section{Introduction}
\label{sec:intro}
Hyperspectral 3D imaging (H3D) has emerged as a powerful technique for object analysis by simultaneously capturing dense hyperspectral reflectance and 3D shape. By providing per-pixel spectral signatures together with geometry, H3D offers richer information than conventional RGB-only or depth-only imaging systems. This capability has been applied to precision agriculture~\cite{dale2013hyperspectral}, mineral exploration~\cite{tripathi2019evaluation}, and cultural heritage digitization~\cite{kim2012developing}, where accurate material and structural analysis is essential.

Building upon the benefits of H3D imaging, broadband hyperspectral 3D imaging (BH3D) represents the next frontier, aiming to achieve both broad spectral coverage and dense sampling within a unified 3D representation. The motivation for BH3D lies in the fact that different wavelength ranges convey complementary physical cues. Visible wavelengths (400--700\,nm) capture fine-grained human-perceptible appearance driven by surface reflectance. Near-infrared (NIR, 700--1000\,nm) wavelengths are sensitive to water content~\cite{okawa2019estimation} and subsurface scattering~\cite{tanaka2013descattering}, enabling, for example, assessment of vegetation health, organic materials, and banknotes~\cite{tang2015high}. Short-wavelength infrared (SWIR, 1000--1500\,nm) further reveals material-specific absorption features, supporting discrimination of minerals, polymers, and biological tissues~\cite{huang2014recent, lu2014medical, sun2024applications, kim2015highly}. Capturing dense spectral measurements across visible, NIR, and SWIR bands therefore enables more comprehensive and reliable material characterization than conventional visible-spectrum H3D imaging. A BH3D system spanning these ranges can jointly support appearance-based perception, subsurface and moisture analysis, and chemical or compositional inference within a unified spectral--geometric representation.

Despite its importance, BH3D remains difficult to achieve with existing systems, especially in terms of broadband spectral imaging. 
While recent hyperspectral imaging systems achieve compactness and accuracy~\cite{baek2017compact,baek2021single,shin2025dense,shi2024learned,yu2025active,li2024inter}, they are typically restricted to narrow spectral bands, operating in either visible, NIR, or SWIR bands. This limitation largely arises from sensor constraints: silicon-based CMOS sensors cover visible–NIR wavelengths, whereas InGaAs sensors are required for SWIR imaging~\cite{hansen2008overview}.

Spectrograph-based methods enable BH3D by combining dispersive optics with scanned slits to achieve high spectral fidelity~\cite{brusco2006system,porter1987system,li20244d}. However, acquiring visible--NIR--SWIR spectra typically requires two different spectrograph modules paired with different image sensors~\cite{li20244d}. This dual-spectrograph configuration substantially increases system complexity, size, and cost, and complicates calibration and synchronization.

In this work, we propose a BH3D imaging system that simultaneously captures accurate 3D geometry and spectral reflectance from 450--1500\,nm, covering the visible to SWIR spectrum. Our key idea is to extend the concept of dispersed structured light~\cite{shin2025dense,shin2024dispersed,yu2025active} to broadband hyperspectral imaging. Instead of using separate spectrographs for visible--NIR and SWIR sensing, we project broadband spectrally encoded structured illumination using a single dispersive element placed in front of a halogen light source. As shown in Figure~\ref{fig:teaser}, the scene is captured by a calibrated stereo pair consisting of a visible–NIR (VNIR) camera and an SWIR camera. To reconstruct the full hyperspectral cube of a static scene, our system employs automatic galvo-mirror scanning to capture images at varying dispersion angles. The projected broadband dispersive illumination provides spectral encoding cues, enabling unified broadband hyperspectral reconstruction, while the stereo setup provides pixel-aligned depth information for accurate geometry estimation and for resolving the wavelength–geometry ambiguity.

Achieving this design introduces challenges. First, the hardware system must precisely combine broadband dispersive illumination with heterogeneous image sensors. Second, accurate modeling of the wavelength-dependent image formation process is required to relate projected patterns to captured measurements across both cameras. Third, a robust reconstruction algorithm is needed to recover dense hyperspectral reflectance and 3D geometry from these coupled observations. 

Our contributions address these challenges through: (1) a BH3D imaging system based on dispersed structured light and visible--NIR and SWIR stereo imaging; (2) an image formation model that captures the spectral--spatial coupling induced by broadband dispersive illumination; and (3) a reconstruction framework that estimates hyperspectral reflectance and depth with high accuracy. We demonstrate that our method achieves broadband hyperspectral reconstruction with a mean spectral angle mapper (SAM) of $0.13$ rad, root mean square error (RMSE) of $0.03$ and mean depth error of 4.5\,mm enabling BH3D imaging across visible to SWIR wavelengths. Using our system, we demonstrate metamerism detection, imaging through opaque layers, capturing hidden patterns on a banknote, and revealing the structure of blood vessels beneath the skin. Code and calibration datas for this paper are at  \href{https://github.com/shshin1210/BH3D}{https://github.com/shshin1210/BH3D}.

\section{Related Work}
\label{sec:related}
\paragraph{Hyperspectral 3D Imaging}
H3D has been developed through various system designs that capture spectral and geometric information. Early approaches combine compressive hyperspectral imaging with external 3D scanners to achieve high spectral and geometric accuracy, but require complex and bulky hardware~\cite{kim20123d}.
More recent work has investigated compact optical designs that encode spectral information through micro- or nano-scale optical elements by modulating the point spread function (PSF), albeit sacrificing spectral accuracy~\cite{baek2021single, shi2024split}. 
Active H3D methods use a conventional RGB projector and camera system where depth is recovered with structured-light patterns and spectral information is inferred from color-coded illumination~\cite{li2019pro, li2022deep}.
However, the limited spectral diversity of the RGB projector limits spectral accuracy. 
Recent advances overcome this limitation by incorporating diffraction grating films in front of RGB projectors to generate densely dispersed hyperspectral structured light~\cite{shin2024dispersed, shin2025dense}.
Despite these advances, most existing approaches remain limited in spectral bandwidth, typically operating only within the visible spectrum.

\begin{figure*}[t]
    \centering
    \includegraphics[width=\linewidth]{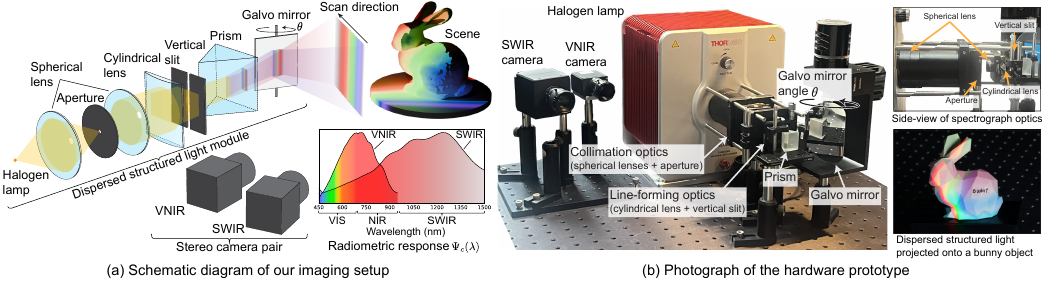}
    \vspace{-3mm}
    \caption{\textbf{BH3D imaging system using dispersed structured light.} (a) The system comprises a broadband dispersed structured light module and a stereo camera consisting of VNIR and SWIR sensors. A halogen lamp, which provides continuous spectral coverage from the visible to the SWIR range, serves as the broadband illumination source. The emitted light is first collimated using a pair of spherical lenses and an aperture, then shaped into a line illumination by a cylindrical lens and a vertical slit. The resulting beam is spectrally dispersed by a prism and subsequently reflected by a galvanometric mirror to illuminate and scan the scene. (b) Photograph of the hardware prototype. Insets show (top-right) a side view of the projection optics/scan head, and (bottom-right) an example dispersed structured-light pattern projected onto a bunny object.
    \vspace{-3mm}
    }
    \label{fig:imaging_system}
\end{figure*}

\paragraph{Hyperspectral Imaging}
Spectrograph-based imaging has enabled broadband hyperspectral acquisition, in pushbroom and whiskbroom architectures. These typically rely on mechanical scanning, such as conveyor-belt translation stages and bulky system configurations~\cite{brusco2006system, porter1987system}. Li et al.~\shortcite{li20244d} employ two cameras with separate optical paths to capture spectrally dispersed measurements across a wide wavelength range. Although this design enables broadband spectral sensing, it introduces substantial system complexity and aberration, requiring multiple optical modules in two different spectrographs. To reduce reliance on mechanical translation, Al-Hourani et al.~\shortcite{al2023line} propose a scanning approach that places a galvo mirror in front of an VNIR camera, enabling angular scanning without physical object motion. However, it remains restricted to the VNIR spectrum. Extending such a system to broadband visible–NIR–SWIR operation would necessitate an additional spectrograph module including a galvo mirror, significantly increasing system complexity. In contrast to these approaches, we achieve accurate BH3D imaging using a single dispersive illumination module.

Snapshot-based methods, such as CASSI~\cite{wagadarikar2008single}, multiplex spatial and spectral information but often face trade-offs between spectral accuracy and hardware form factor. Recent efforts have sought to improve snapshot methods through metasurfaces~\cite{li2024inter}, learned filters with dual-pixel sensors~\cite{shi2024learned}. Nevertheless, these methods remain limited by narrow spectral bands.

\paragraph{Hyperspectral Structured Light}
Structured light projects illumination patterns onto a scene to recover scene properties by analyzing the captured images~\cite{geng2011structured,ni2020metasurface,baek2021polka}. 
Beyond geometry estimation, it has also been extended to probe richer scene attributes such as light transport using polarization-aware, spatio-temporal illumination as well as spectral reflectance~\cite{ichikawa2024spiders,baek2021polarimetric,o20143d}.
Early approaches achieved spectral modulation using dispersive optics such as prisms or gratings together with attenuation masks~\cite{mohan2008agile,hostettler2015dispersion}. More flexible systems combine dispersive elements with digital micro-mirror devices to control both spatial and spectral dimensions, enabling high-dimensional modulation at the cost of increased optical complexity and calibration effort~\cite{rice2006development,rice2007hyperspectral,xu2020hyperspectral}.
Recent systems combine dispersive optics with conventional projectors to emit spatially multiplexed patterns for H3D imaging~\cite{shin2024dispersed,shin2025dense}, or employ fixed patterns for hyperspectral imaging with direct/global separation~\cite{ishihara2025spatio}. Yu et al.~\shortcite{yu2025active} employ a diffraction grating and polygon mirror to scan dispersed illumination by mirror rotation. Yet, existing hyperspectral structured light systems are often limited in spectral ranges or suffer from low signal-to-noise ratios. We address these limitations by using with a broadband halogen source and a prism–galvo configuration, combined with a VNIR-SWIR stereo pair. Our image formation model and reconstruction method exploit this system design to reconstruct hyperspectral reflectance spanning the visible to SWIR range.

\section{Broadband Dispersed Structured Light}
\label{sec:method}
\subsection{Imaging System}
\paragraph{System Design}
We design an imaging system that jointly captures depth and broadband hyperspectral reflectance from the visible to the SWIR range (Figure~\ref{fig:imaging_system}). The system uses a visible-to-SWIR broadband illumination module that projects dispersed structured light, enabling BH3D imaging with a single spectrograph. The illumination module employs a halogen light source to provide continuous broadband emission, which is collimated using plano-convex spherical lenses and shaped into a thin vertical line by a cylindrical lens and slit. A broadband dispersive prism introduces wavelength-dependent spatial dispersion. The dispersed structured light is laterally swept across the scene using a galvo mirror to enable automated scanning. For image acquisition, we employ a stereo camera setup consisting of a VNIR and a SWIR camera to enable simultaneous capture over the visible–SWIR spectrum. Although the SWIR sensor nominally covers the visible range, there exists a substantial signal imbalance between visible and SWIR bands. We therefore employ a dedicated VNIR camera, which is sensitive to the visible–NIR range, as detailed in the Supplementary Document.
In addition, to handle the large wavelength-dependent variation in illumination strength and sensor sensitivity, we acquire multiple measurements at different exposure settings and fuse them into HDR images.

\begin{figure}[t]
    \centering 
    \includegraphics[width=\linewidth]{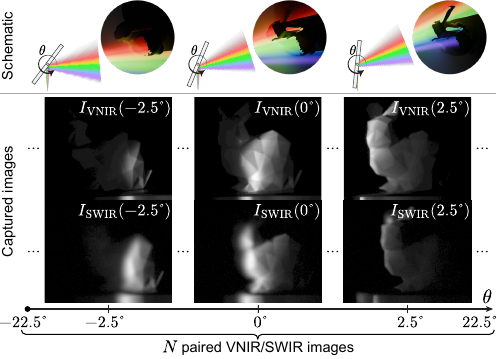}
    \caption{\textbf{Image acquisition process with galvo-scanned dispersed structured light.} Top: Schematic illustration of a galvo mirror rotated in discrete steps, parameterized by the scan angle $\theta$, to sweep the dispersed structured-light pattern across the scene. Middle/Bottom: Example VNIR and SWIR images captured by our prototype at different galvo mirror angles. We acquire $N$ paired VNIR/SWIR images over the scan range $\theta \in [-22.5^\circ,\,22.5^\circ]$.}
    \label{fig:galvo_rotation}
\end{figure}

\paragraph{Principle of Operation}
Our system acquires broadband hyperspectral and geometric information through scanning of the dispersed structured light. Specifically, we incrementally rotate the galvo mirror to laterally sweep the broadband dispersed structured light across the field of view. This scanning mechanism ensures that every spatial point in the scene is sequentially illuminated by the full range of broadband wavelengths. Consequently, as shown in Figure~\ref{fig:galvo_rotation}, we obtain VNIR/SWIR image pairs over galvo angles $\theta \in [-22.5^\circ, 22.5^\circ]$. This captured sequence encodes the wavelength-dependent reflectance and the 3D geometry of the scene. The scanning procedure is shown in the Supplementary Video.

\subsection{Image Formation}
At a galvo mirror angle $\theta$, broadband dispersed structured light is projected onto a scene, and the reflected light is captured by a camera $c \in \{\text{VNIR}, \text{SWIR}\}$, which integrates the incoming radiance according to its spectral sensitivity.
Accordingly, we model the captured intensity at pixel $(x,y)$ as:
\begin{equation}
I_c(x,y,\theta) \;=\; \int \Omega_c(\lambda)\, H_c(x,y,\lambda)\, \Phi_c(x,y,Z_c,\theta,\lambda)\, d\lambda ,
\label{eq:image_formation}
\end{equation}
where $\lambda$ is the wavelength, $\Omega_c(\lambda)$ denotes the spectral sensitivity of camera $c$, $H_c(x,y,\lambda)$ is the hyperspectral reflectance at the camera view $c$, and $\Phi_c(x,y,Z_c,\theta,\lambda)$ is the broadband dispersed structured light projected onto the scene.
Here, $Z_c$ is the depth of the corresponding scene point measured from the optical center of camera~$c$.

\begin{figure}[t]
    \includegraphics[width=\linewidth]{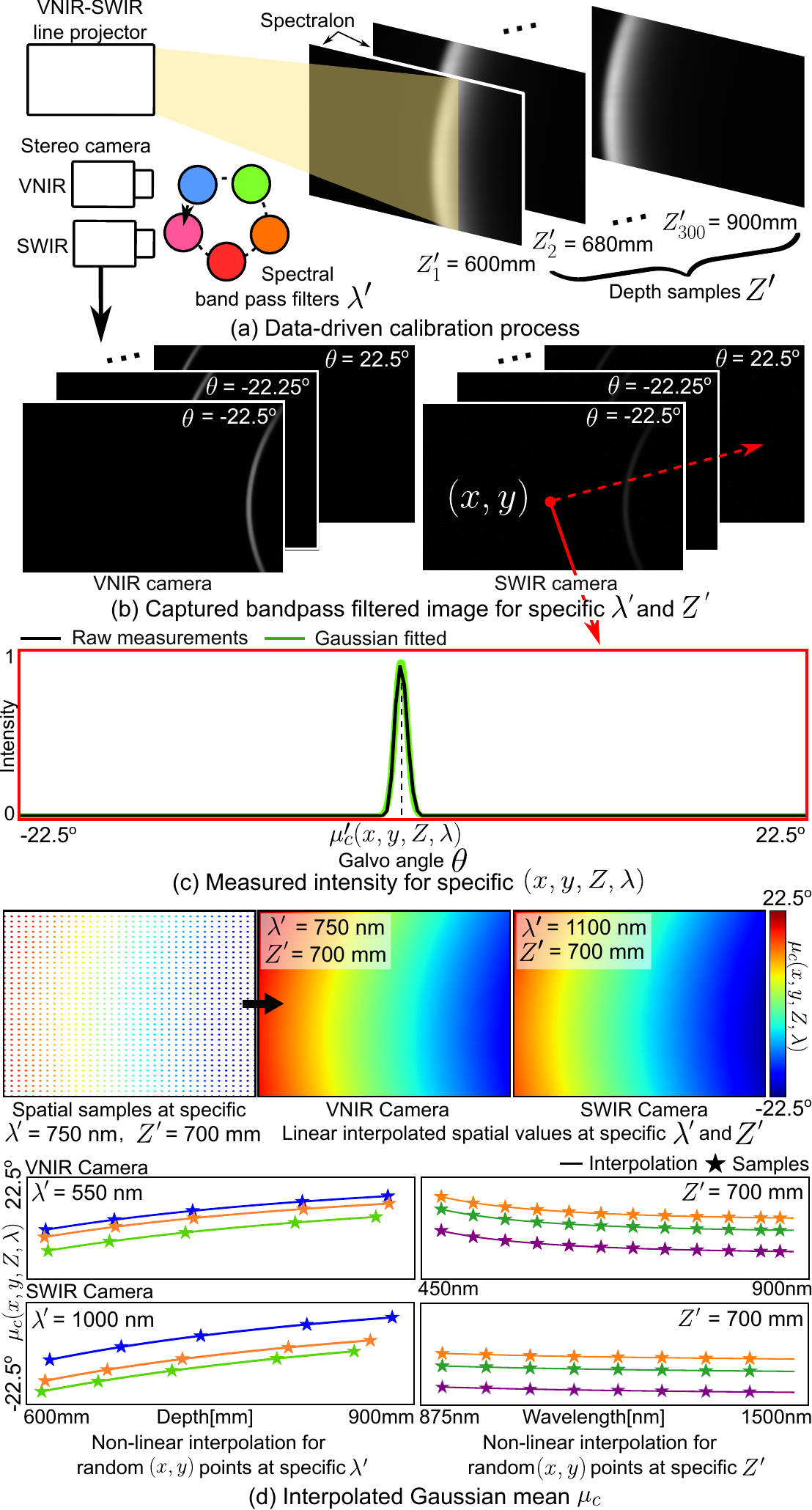}
    \vspace{-7mm}
    \caption{\textbf{Data-driven method for Gaussian model parameters.}
    (a) Calibration procedure, in which a broadband dispersed line is swept across a Spectralon target using narrow band-pass filters mounted on each camera.
    (b) Captured images at a given wavelength $\lambda'$ and depth $Z'$. (c) Intensity at a pixel $(x,y)$ as a function of the galvo mirror angle, showing the raw and fitted versions. (d) Interpolation of the Gaussian mean $\mu_c$ across spatial, depth, and wavelength dimensions.}
    \vspace{-5mm}
    \label{fig:dispersive_weight_model}
\end{figure}

\paragraph{Gaussian Model for Structured Light}
The key component of our image formation model is the term $\Phi_c(x,y,Z_c,\theta,\lambda)$, which describes the intensity contribution of wavelength $\lambda$ at galvo mirror angle $\theta$ to the scene point corresponding to the pixel $(x,y)$ at depth $Z_c$.

In an idealized setup, for a specific wavelength $\lambda$, the corresponding projected line at a spatial position would be confined to a single-pixel-wide line. In practice, however, the projected spectral pattern exhibits a blurry spatial profile as shown in Figure~\ref{fig:dispersive_weight_model}(c). This is because of the finite diameter of the collimated beam, which prevents perfect point focusing through the cylindrical lens, and diffraction effects arising from the finite apertures, and aberration within the optical path. This inherent blurring means that each point in the scene integrates a narrow band of wavelengths at each galvo position. 
We model this behavior using a Gaussian function defined over the galvo mirror angle $\theta$:
\begin{equation}
\Phi_c(x,y,Z_c,\theta,\lambda)
\;=\;
\exp\!\left(
-\frac{\big(\theta - \mu_c(x,y,Z_c,\lambda)\big)^2}
{2\,\sigma_c^2(x,y,Z_c,\lambda)}
\right)
\, T(\lambda)\, R(\lambda)\, \frac{E(\lambda)}{Z_c^2},
\label{eq:dispersed_structured_light}
\end{equation}
where $T(\lambda)$ denotes the wavelength-dependent transmittance of the optical elements, $R(\lambda)$ is the reflectance of the galvo mirror, and $E(\lambda)$ is the emission intensity of the halogen light source. Assuming that the cameras and light source are much closer to each other than to the scene, $Z_c$ is used as the distance for the inverse square law.

The Gaussian mean $\mu_c(x,y,Z_c,\lambda)$ corresponds to the galvo mirror angle at which wavelength $\lambda$ contributes the maximum intensity for a given pixel and depth, while the standard deviation $\sigma_c(x,y,Z_c,\lambda)$ represents the blur of the dispersed structured light line.
Equation~\eqref{eq:dispersed_structured_light} captures the effect that, as the galvo mirror sweeps the dispersed light across the scene, each wavelength is observed at slightly different galvo angles and contributes smoothly across neighboring angles, rather than appearing at a single discrete location. 

\vspace{-2mm}
\subsection{Calibration}
\label{subsec:calibration}
\paragraph{Gaussian Model Parameters}
We estimate the Gaussian parameters $\mu_c(x,y,Z,\lambda)$ and $\sigma_c(x,y,Z,\lambda)$ of our image formation model through a data-driven calibration procedure.
As shown in Figure~\ref{fig:dispersive_weight_model}(a,b), broadband dispersed structured light is swept across a Spectralon planar target, placed at multiple depth samples $Z'$ by rotating the galvo mirror from $-22.5^\circ$ to $22.5^\circ$ in $0.25^\circ$ increments. This procedure ensures that all spatial locations are illuminated over a range of depths.
Narrow band-pass filters are placed in front of both stereo cameras to isolate wavelength-specific responses at discrete wavelength samples $\lambda'$, selected within the respective reconstruction ranges of each VNIR and SWIR cameras.

For each band-pass filter, we record intensity versus galvo angle $\theta$ to obtain an angular response profile as Figure~\ref{fig:dispersive_weight_model}(c), which is subsequently modeled using a Gaussian function.
The peak location of the fitted curve provides the estimated mean angle $\mu_c'$, while the angular spread yields the standard deviation $\sigma_c'$.
These fitted values constitute discrete samples at known spatial locations, depths, and wavelengths.

Given the discrete estimates $\mu_c'$ and $\sigma_c'$, we interpolate them to obtain continuous parameter fields $\mu_c(x,y,Z,\lambda)$ and $\sigma_c(x,y,Z,\lambda)$ over the full spatial, depth, and wavelength domains.
We use linear interpolation along the spatial dimensions $(x,y)$ and nonlinear interpolation along the depth $Z$ and wavelength $\lambda$ dimensions to capture the smooth yet nonlinear variations.
This process yields dense parameter fields for $\mu_c$ and $\sigma_c$, and we evaluate the Gaussian model throughout the entire imaging volume, as shown in Figure~\ref{fig:dispersive_weight_model}(d). Details are provided in the Supplementary Document.

\vspace{-2mm}
\paragraph{Geometric and Radiometric Parameters}
We estimate the intrinsic and extrinsic parameters of the stereo cameras using a checkerboard-based calibration method~\cite{zhang2002flexible}. The radiometric response is modeled as $\Psi_c(\lambda) = \Omega_c(\lambda) T(\lambda) R(\lambda) E(\lambda)$ and optimized by projecting a broadband dispersed line over the full galvo scan onto a $99\%$ reflectance Spectralon target. Given pre-calibrated Gaussian parameters, $\Psi_c(\lambda)$ is estimated by minimizing the $l_1$ residual between captured and rendered intensities across all scan angles. Refer to Supplementary Documents for more details. Figure~\ref{fig:imaging_system}(a) shows the resulting optimized radiometric parameters for each camera.

\begin{figure*}
    \includegraphics[width=\linewidth]{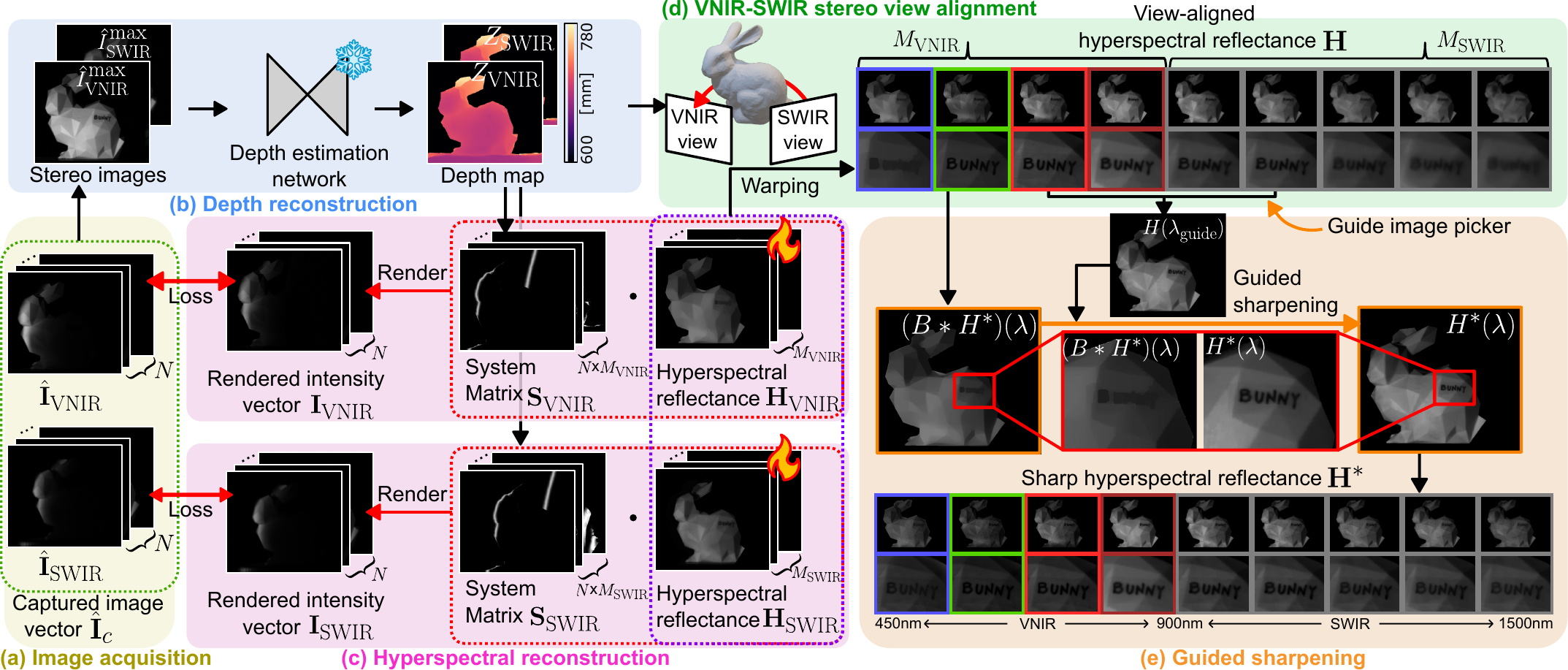}
    \caption{
    \textbf{Hyperspectral 3D reconstruction process.} 
    (a) Images are captured under $N$ number of galvo mirror angles for each VNIR and SWIR cameras.
    (b) We synthesize stereo images using captured image vectors, then reconstruct accurate depth map with a pre-trained depth estimation network~\cite{wen2025foundationstereo}.
    (c) Hyperspectral reflectance is recovered by minimizing the photometric residual between the captured image vectors and the rendered intensity vectors. In this analysis-by-synthesis framework, the reconstructed depth maps are utilized to formulate the system matrix within the image formation model.
    (d) The VNIR and SWIR hyperspectral reflectance maps are aligned to a common reference view (VNIR) by warping them using the estimated depth. Here we earn the view-aligned hyperspectral reflectance $\textbf{H}$.
    (e) To compensate for spectral-dependent focus blur, sharp image is selected as a guide image for guided sharpening. Finally, we get a view aligned and sharp VNIR-SWIR hyperspectral reflectance $\textbf{H}^*$ spanning wavelength from 450\,nm to 1500\,nm.}
    \label{fig:recon}
\end{figure*}

\section{Hyperspectral 3D Reconstruction}
\label{sec:reconstruction}

\subsection{Depth Reconstruction}
\label{sec:depth_recon}
For depth reconstruction, we synthesize stereo images by taking a maximum-intensity projection of the measurements across the scan-angle direction, as illustrated in Figure~\ref{fig:recon}(a). Specifically, given $N$ captured images $\{\hat{I}_c(\theta_i)\}^N_{i=1}$ captured at different galvo mirror angles, we synthesize a stereo image by selecting the maximum\footnote{While alternative aggregation strategies (e.g., averaging) are possible, we adopt max aggregation because the peak response produces a higher-SNR stereo image and reduces the accumulation of additive sensor noise, which is particularly pronounced in our SWIR sensor. We provide comparisons of aggregation strategies in the supplementary material.} intensity at each spatial location $(x,y)$ across all angles:
\begin{equation}
\hat{I}^\text{max}_c(x,y)
=
\max_{i \in \{1,\ldots,N\}} \; \hat{I}_c(x,y,\theta_i),
\label{eq:create_stereo}
\end{equation}
where $c\in\{\text{VNIR}, \text{SWIR}\}$ denotes the two stereo views corresponding to the left and right cameras, respectively. The resulting $\hat{I}^\text{max}_c$ serves as a stereo pair of consistent reference for geometric matching and enables robust disparity estimation. 

\paragraph{Cross-modal Stereo Matching}
Given pre-calibrated stereo parameters, we rectify the input images and feed them into the pretrained FoundationStereo network~\cite{wen2025foundationstereo} to estimate dense disparity maps. Although VNIR and SWIR images exhibit notable appearance and intensity discrepancies, FoundationStereo remains robust to such cross-spectral variations by leveraging learned geometric and structural features rather than relying on raw intensity consistency. We first verify this behavior using a controlled RGB stereo experiment, where different color channels are assigned to each view to introduce significant intensity inconsistency. Despite this mismatch, the resulting depth remains nearly identical to that obtained from grayscale images, indicating that intensity differences alone do not degrade matching performance.

We further evaluate a more challenging case using spectrally distant bands at 550\,nm and 1150\,nm from the scene in Figure~\ref{fig:vinyl}, which exhibit clear semantic differences. Nevertheless, the estimated depth closely matches the reference with low error, as the visible outer surface and the internal structure share similar depth values, resulting in only minor geometric discrepancies between the two views. The estimated disparities are then converted to depth using the calibrated stereo geometry and warped back to the original VNIR and SWIR image coordinates, as shown in Figure~\ref{fig:recon}(b). Additional discussion on more extreme cross-spectral cases is provided in the Supplementary Document.

\subsection{Hyperspectral Reconstruction}
\label{sec:rendering-based-hyperspectral-recon}
We perform hyperspectral reconstruction using an analysis-by-synthesis framework that incorporates a geometry-dependent image formation model. The reconstructed depth maps serve as geometric priors in this formulation, as shown in Figure~\ref{fig:recon}(c).

\paragraph{Vectorized Image Formation}
We first vectorize the image formation model of Equation~\eqref{eq:image_formation} along the galvo mirror angle $\theta$ dimension for each pixel $(x,y)$ of the camera $c$ as:
\begin{equation}
\mathbf{I}_c(x,y) = \mathbf{S}_c(x,y) \mathbf{H}_c(x,y),
\label{eq:matrix}
\end{equation}
where $\mathbf{I}_c(x,y) \in \mathbb{R}^{N \times 1}$ denote the intensity vector corresponding to $N$ galvo mirror angles.
The image formation process for $\mathbf{I}_c(x,y)$ can then be expressed as the multiplication of a system matrix $\mathbf{S}_c(x,y) \in \mathbb{R}^{N \times M_c}$ and the hyperspectral reflectance vector $\mathbf{H}_c(x,y) \in \mathbb{R}^{M_c \times 1}$, where $M_c$ is the number of discrete wavelength samples for each camera $c$. 
Here, $\mathbf{S}_c(x,y)$ encodes both the Gaussian model of the dispersed structured light and the camera spectral sensitivity $\Omega_c$:
\begin{equation}
\setlength{\arraycolsep}{3pt}
\resizebox{\linewidth}{!}{
$\begin{aligned}
\mathbf{I}_c(x,y)
&=
\begin{bmatrix}
I_c(x,y,\theta_1) \\
\vdots \\
I_c(x,y,\theta_N)
\end{bmatrix},
\qquad
\mathbf{H}_c(x,y)
=
\begin{bmatrix}
H_c(x,y,\lambda_1) \\
\vdots \\
H_c(x,y,\lambda_{M_c})
\end{bmatrix}, \\[6pt]
\mathbf{S}_c(x,y)
&=
\begin{bmatrix}
\Omega_c(\lambda_1)\Phi_c(x,y,Z_c,\theta_1,\lambda_1)
& \!\dots\! &
\Omega_c(\lambda_M)\Phi_c(x,y,Z_c,\theta_1,\lambda_{M_c}) \\
\vdots & \ddots & \vdots \\
\Omega_c(\lambda_1)\Phi_c(x,y,Z_c,\theta_N,\lambda_1)
& \!\dots\! &
\Omega_c(\lambda_M)\Phi_c(x,y,Z_c,\theta_N,\lambda_{M_c})
\end{bmatrix}
.
\end{aligned}$
}
\label{eq:matrix_explanation}
\end{equation}

\paragraph{Per-camera Spectral Resolution and Bandwidth}
In our implementation, the VNIR camera reconstructs hyperspectral refletance vecotr $\textbf{H}_\text{VNIR}$ over 450\,nm to 900\,nm at a 20\,nm intervals, while the SWIR camera reconstructs $\textbf{H}_\text{SWIR}$ over 875\,nm to 1500\,nm at 25\,nm intervals.

\paragraph{Optimization}
Using the matrix–vector formulation of Equation~\eqref{eq:matrix}, we reconstruct each hyperspectral reflectance image $\mathbf{H}_c$ by solving the following optimization problem:
\begin{equation}
\underset{\mathbf{H}_c}{\mathrm{argmin}}
\;\|\mathbf{S}_c \mathbf{H}_c - \hat{\mathbf{I}}_c\|_2^2
+ \left(
\left\| \nabla_\lambda \mathbf{H}_c \;\oslash\; \Psi_c \right\|_1
+ \left\| \nabla_{xy} \mathbf{H}_c \;\oslash\; \Psi_c \right\|_1
\right),
\label{eq:reconstruction_equation}
\end{equation}
where $\hat{\mathbf{I}}_c$ indicates the captured image vector and $\oslash$ denotes element-wise division.
The first term measures the reconstruction loss between rendered and captured vectors, while the second term enforces sparsity in spectral and spatial gradients, weighted by the inverse wavelength-dependent radiometric response parameter $\Psi_c$. This adaptive weighting reflects wavelength-dependent signal variations, encouraging stronger regularization in low-signal bands and weaker regularization in high-signal bands to preserve fine structures. We optimize using Adam~\cite{kingma2015adam} with a learning rate of 0.1.

\begin{figure*}
    \centering
        \includegraphics[width=\linewidth]{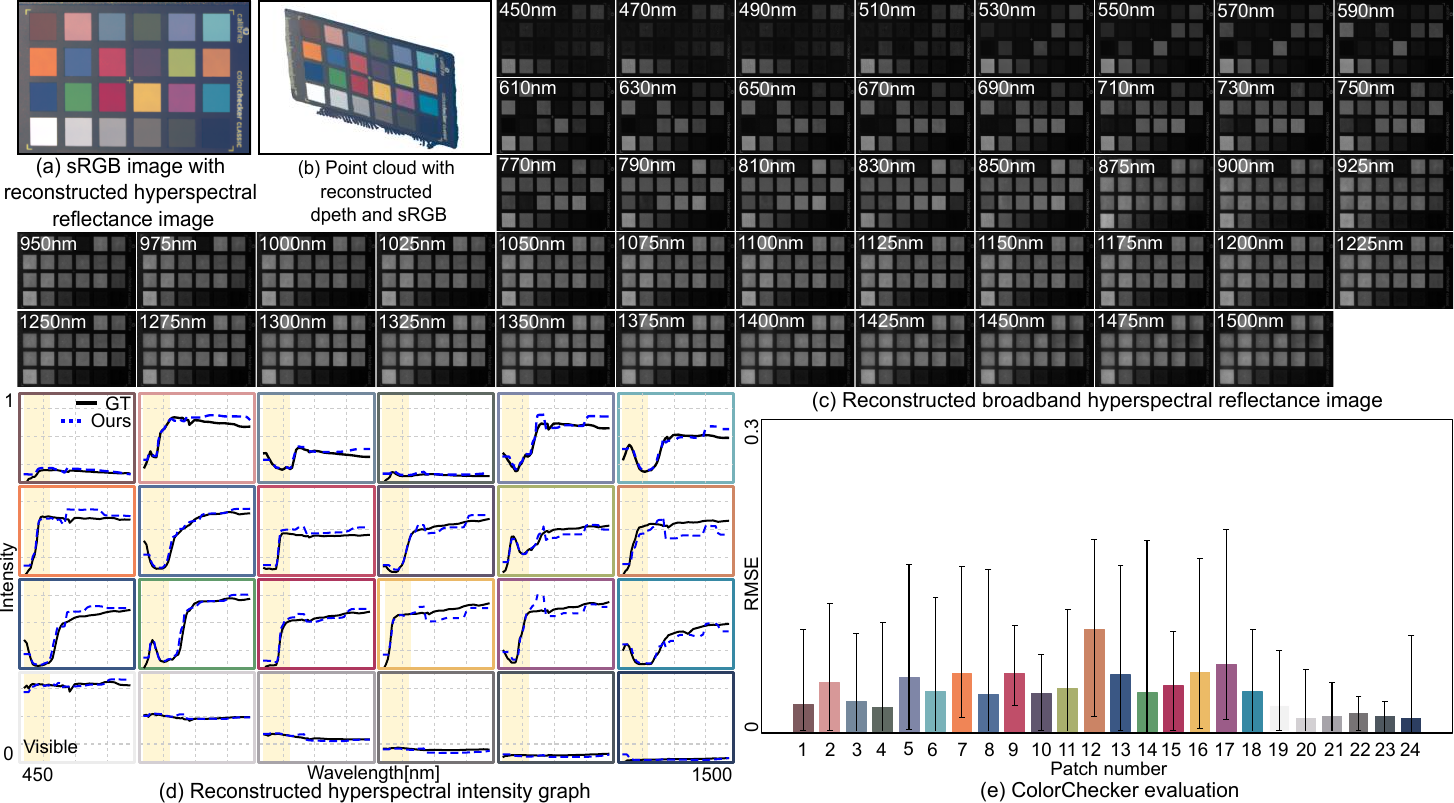}
        \vspace{-6mm}
        \caption{\textbf{ColorChecker evaluation.} Our method obtains a mean {SAM} of 0.13 rad and a mean {RMSE} of 0.03 across 24 patches, reflecting high spectral accuracy. (a) sRGB image with reconstructed hyperspectral reflectance. (b) Reconstructed depth. (c) Reconstructed broadband hyperspectral reflectance images across the VNIR–SWIR spectrum. (d) Reconstructed hyperspectral reflectance curves for all 24 ColorChecker patches. (e) Patch-wise RMS error and error bar.}
        \label{fig:colorchecker}
\end{figure*}

\paragraph{VNIR-SWIR Stereo View Alignment}
We fuse the reconstructed hyperspectral reflectance volumes $\mathbf{H}_\text{VNIR}$ and $\mathbf{H}_\text{SWIR}$ into a single spectral volume $\mathbf{H}$ spanning 450–1500\,nm with a total of $47$ spectral bands, using geometry-guided remapping. VNIR pixels are unprojected to 3D using the reconstructed depth map $Z_\text{VNIR}$, transformed into the SWIR camera coordinate frame, and reprojected onto the SWIR image plane to establish pixel-wise correspondences. To suppress ghosting artifacts, correspondences with large depth discrepancies between the VNIR and SWIR depth maps are discarded. Results are shown in Figure~\ref{fig:recon}(d).

\begin{figure}[h]
    \centering
            \includegraphics[width=\linewidth]{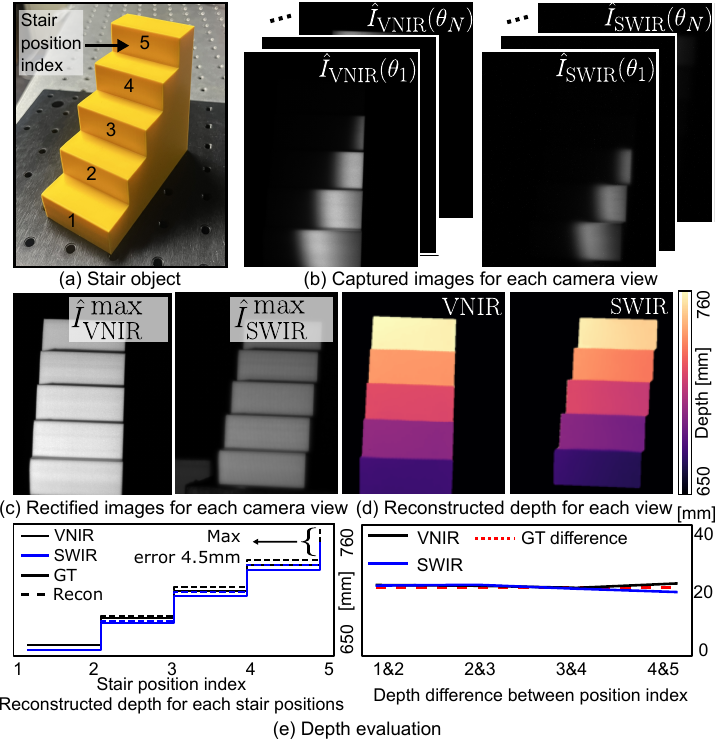}
            \vspace{-4mm}
            \captionof{figure}{
                \textbf{Depth evaluation.} (a) 3D-printed stair case object. (b) Captured images for each camera view under broadband dispersed structured light. (c) Synthesized stereo images for each camera view. (d) Reconstructed depth and (e) center-column depth evaluation.
            }
    \label{fig:depth_eval}
\end{figure}

\begin{figure}
    \centering
        \includegraphics[width=\linewidth]{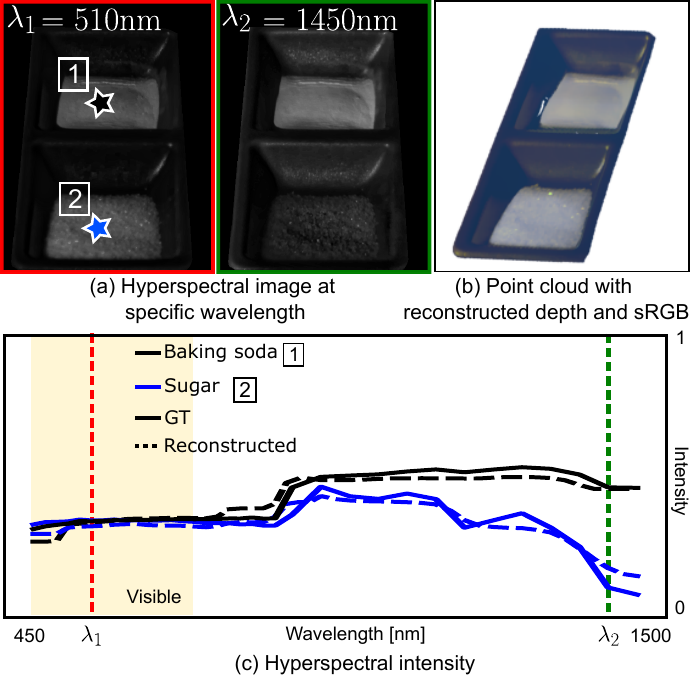}
        \vspace{-6mm}
        \captionof{figure}{\textbf{Baking soda vs. sugar.} We demonstrate that baking soda and sugar exhibit nearly similar spectral intensities in the visible wavelength range, while showing a clear spectral difference in SWIR region. (a) 510\,nm and 1450\,nm images. (b) Point cloud with reconstructed depth and sRGB and (c) Intensity graph for each baking soda and sugar.}
    \label{fig:powder}
\end{figure}

\begin{figure}[h]
    \centering
        \includegraphics[width=\linewidth]{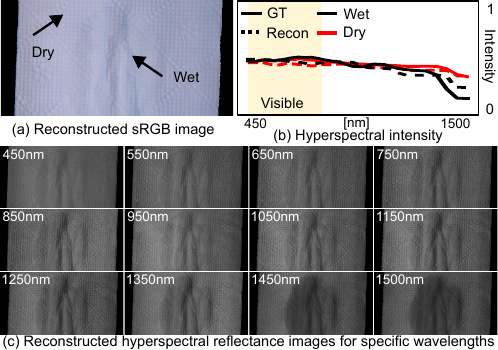}
        \vspace{-4mm}
        \captionof{figure}{\textbf{Wet vs. dry materials.} Water exhibits strong absorption in the SWIR region, resulting in low reflectance at longer wavelengths. (a) sRGB with hyperspectral image. (b) Intensity graph of wet and dry material. (c) Broadband reflectance images.}
        \label{fig:water}
\end{figure}
    
\paragraph{Guided Sharpening.}
While we reconstruct hyperspectral reflectance $\mathbf{H}$, certain wavelength bands exhibit spatial blur $B(\lambda)$ (Figure~\ref{fig:recon}(d)) due to wavelength-dependent focal length variations across the VNIR–SWIR range. We model this effect as $H(\lambda) = (B * H^*)(\lambda)$, where $H^*(\lambda)$ denotes the latent sharp hyperspectral image.

To mitigate this blur, we repurpose the guided sharpening technique used in Baek et al.~\shortcite{baek2017compact} that applied guided filtering~\cite{he2012guided} to the entire hyperspectral channels caused by reconstruction artifacts.
Different from the previous work, we use guided sharpening to mitigate optics-induced blur in partial hyperspectral channels by exploiting the availability of relatively sharp hyperspectral channels, called \textit{guide} wavelengths lying in the depth-of-field. Specifically, for each wavelength $\lambda$ channel, we pick the nearest guide wavelength $\lambda_{\text{guide}}$ from the sets $\Lambda_{\text{VNIR}} = \{ \lambda \mid 510 \le \lambda \le 850\,\text{nm} \}$ and $\Lambda_{\text{SWIR}} = \{ \lambda \mid 950 \le \lambda \le 1200\,\text{nm} \}$, and use the corresponding sharp hyperspectral image $H(\lambda_{\text{guide}})$ as the guide.
Guided sharpening~\cite{he2012guided} transfers high-frequency spatial details from a guide while preserving spectral characteristics, and is applied separately to the VNIR and SWIR ranges to produce sharper hyperspectral images across the full spectrum as Figure~\ref{fig:recon}(e). Additional details are provided in the Supplementary Document.

\section{Assessment}
\label{sec:assessment}
\subsection{Hyperspectral Depth Accuracy}

\paragraph{Hyperspectral Reflectance of a ColorChecker}
We first evaluate hyperspectral reconstruction accuracy using a ColorChecker chart. Figure~\ref{fig:colorchecker}(a) shows the sRGB image using the reconstructed hyperspectral reflectance image shown in Figure~\ref{fig:colorchecker}(c). Figure~\ref{fig:colorchecker}(d,e) shows that the reconstructed spectral reflectance curves over the wavelength range of 450-1500\,nm closely match with the ground-truth measurements. 
In addition, we compute the spectral angle mapper (SAM) values between the reconstructed and ground-truth spectra, achieving a mean SAM of 0.13 radians across all patches, indicating high spectral reconstruction accuracy.

\paragraph{Hyperspectral Reflectance on Real-world Scenes}
For all reconstruction results, we compare the reconstructed hyperspectral reflectance at selected spatial locations with corresponding ground-truth measurements, showing close agreement. Ground truth is obtained using a broadband VNIR–SWIR halogen light source (Thorlabs SLS201) and narrow spectral bandpass filters, with pre-calibrated illumination and camera spectral responses. We detail this process in the Supplementary Document.

\paragraph{Depth of a Known Shape}
We evaluate depth reconstruction accuracy using a 3D-printed staircase target with five discrete depth levels, each separated by a known height difference of 20\,mm. Figure~\ref{fig:depth_eval} shows the captured images under broadband dispersed structured-light patterns, the corresponding synthesized stereo images $\hat{I}_c^{\text{max}}$, and the reconstructed depth map. Our method achieves a mean absolute depth error of 4.5\,mm, demonstrating high depth accuracy.

\vspace{-2mm}
\subsection{Hyperspectral 3D Reconstruction}
We demonstrate our BH3D results on a diverse set of objects. Complete broadband hyperspectral reflectance images covering the full wavelength range are included in the Supplementary Document and Supplementary Video.

\vspace{-1mm}
\paragraph{Baking Soda vs. Sugar}
Figure~\ref{fig:powder} shows that baking soda and sugar appear as similarly white powders under visible illumination and exhibit similar reflectance across visible wavelengths. However, at 1450\,nm in the SWIR band, sugar shows significantly lower reflectance than baking soda, enabling reliable discrimination between the two materials. This material-specific contrast is particularly valuable for applications such as pharmaceutical manufacturing and food processing, where accurate identification of visually similar compounds is critical for quality control and safety assurance.
\vspace{-1mm}
\paragraph{Wet vs. Dry Materials}
Water exhibits strong absorption at longer wavelengths, causing reflectance to decrease in the NIR and attenuate further in the SWIR~\cite{okawa2019estimation}. As shown in Figure~\ref{fig:water}, materials saturated with water exhibit a drop in reflectance beyond the visible band that remains low across the NIR–SWIR range whereas the dry region remains relatively still. This distinctive spectral behavior enables robust discrimination between wet and dry regions even when visual cues are ambiguous.

\begin{figure}
    \centering
        \includegraphics[width=\linewidth]{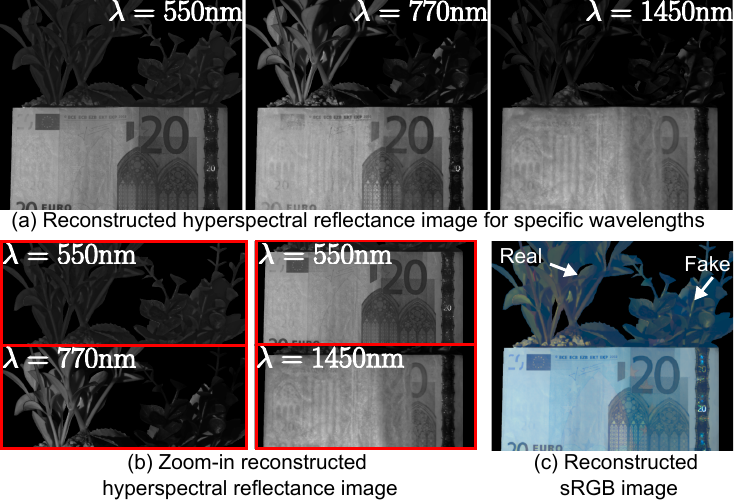}
        \vspace{-7mm}
        \captionof{figure}{
            \textbf{Fake vs. real plants and hidden patterns on a bill.} We highlight spectral difference between real and fake plants and expose hidden banknote security patterns in SWIR. Real plants exhibit elevated NIR–SWIR reflectance, whereas artificial plants remain low, and visible-range security features become indistinct in SWIR. (a) Reconstructed hyperspectral reflectance image for specific wavelengths. (b) Zoom-in reconstructed hyperspectral reflectance image. (c) sRGB image with reconstructed hyperspectral image. 
        }
        \label{fig:money}
\end{figure}

\begin{figure}
    \centering
        \includegraphics[width=\linewidth]{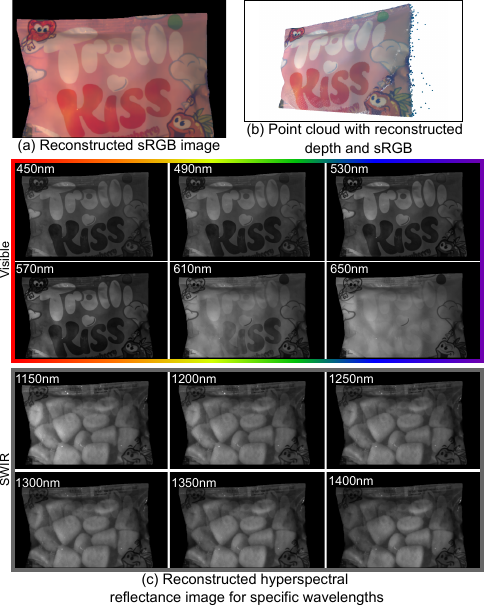}
        \vspace{-7mm}
        \captionof{figure}{
            \textbf{Seeing through opaque vinyl.} The jelly packaging film blocks visible wavelengths but allows transmission of longer wavelengths in the NIR-SWIR region. (a) sRGB with reconstructed hyperspectral image. (b) Point cloud with reconstructed depth and sRGB. (c) Specific broadband hyperspectral images.
        }
        \label{fig:vinyl}
\end{figure}

\vspace{-2mm}
\paragraph{Fake vs. Real Plants and Hidden Patterns on a Bill}
Figure~\ref{fig:teaser} and ~\ref{fig:money} show a scene containing both real and fake plants, with a \texteuro20 banknote placed in the foreground. 
Reconstructed hyperspectral reflectance images in the visible and SWIR ranges reveal distinct material-dependent spectral behaviors.
While the plants appear similar in the visible domain, their spectra differ in the NIR–SWIR range, with the real plant exhibiting higher reflectance. The banknote shows wavelength-dependent appearance changes, where fine texts and security patterns visible in the visible range become invisible in SWIR. The depth-based point cloud captures complex 3D structures, including thin leaves and overlapping geometries.

\begin{figure}
    \centering
        \includegraphics[width=\linewidth]{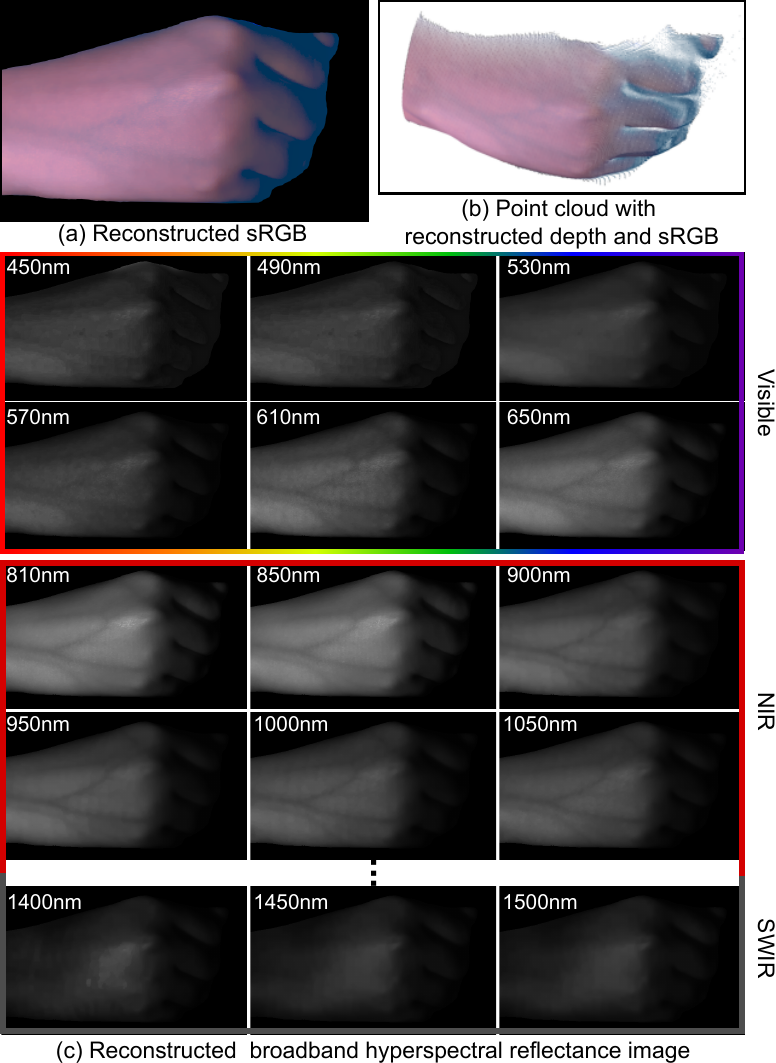}
        \vspace{-7mm}
        \captionof{figure}{
            \textbf{Blood vessels.} (a) sRGB image with our reconstructed hyperspectral reflectance image for visible wavelengths. (b) Point cloud with reconstructed depth and sRGB. (c) Reconstructed specific broadband hyperspectral reflectance image.
        }
        \label{fig:hand}
\end{figure}

\vspace{-1mm}
\paragraph{Seeing Through Opaque Vinyl}
Figure~\ref{fig:vinyl} shows that the vinyl packaging material blocks visible light and prevents visual inspection by the human eye, whereas in the SWIR band it becomes sufficiently transparent to reveal structures behind it. This wavelength-dependent transmissivity enables non-destructive inspection scenarios in which visual access through packaging materials is required.

\vspace{-1mm}
\paragraph{Blood Vessels}
Figure~\ref{fig:hand} shows reconstructed hyperspectral reflectance images of a hand. In the NIR range, subcutaneous veins are visible due to deeper light penetration. As the wavelength increases into the SWIR range, contrast gradually decreases because of stronger absorption, which limits penetration depth and causes vascular structures to disappear. 

\section{Discussion}
\label{sec:discussion}

\paragraph{Broadband Achromatic Imaging}
A key challenge in broadband imaging is wavelength-dependent optical aberration, particularly chromatic focal length variation. We mitigate the resulting blur using guided sharpening, which is effective when image structure is consistent across wavelengths but may introduce inaccuracies otherwise. Future work could address this limitation using apochromatic optics or physically grounded models that explicitly account for wavelength-dependent PSFs to improve spatial fidelity.

\paragraph{Acquisition Speed} 
One limitation of our prototype is acquisition speed: capturing a full hyperspectral 3D volume takes approximately 4 minutes due to galvanometric scanning and HDR imaging. Future work could extend our approach to dynamic scenes by exploring end-to-end differentiable imaging frameworks that jointly optimize coded illumination and reconstruction for sparse acquisition.

\section{Conclusion}
\label{sec:conclusion}

This paper has presented a BH3D imaging method that reconstructs accurate geometry and dense broadband spectral reflectance using a single spectrograph. By generalizing dispersed structured light to the broadband regime, our approach overcomes the hardware complexities of multi-spectrograph systems and unifies VNIR and SWIR observations through a Gaussian image formation model and a joint reconstruction method.
As demonstrated in our experiments, our system captures intrinsic material properties that are inaccessible to conventional RGB-D sensors, enabling applications including seeing through opaque vinyl, discriminating wet and dry surfaces, and verifying material authenticity. 
We believe this work serves as a step toward making broadband high-dimensional physical sensing more accessible for visual computing and beyond.

\begin{acks}
This work was supported by the National Research Foundation of Korea (NRF) grants funded by the Korea government (MSIT) (Nos. RS-2024-00438532, RS-2023-00211658) and the Ministry of Education through the Basic Science Research Program (No. 2022R1A6A1\\A03052954), as well as the Institute of Information \& Communications Technology Planning \& Evaluation (IITP) grants funded by the Korea government (MSIT) (No. RS-2024-0045788, No. RS-2019-II191906 for the Artificial Intelligence Graduate School Program at POSTECH, and No. IITP-2026-RS-2024-00437866 for the ITRC).
\end{acks}


\bibliographystyle{ACM-Reference-Format}
\bibliography{Source/references}

\end{document}